\documentclass[runningheads]{llncs}
\usepackage[T1]{fontenc}
\usepackage{amsmath}
\usepackage{graphicx}
\usepackage{multirow}
\usepackage{multicol}

\begin{document}
\title{Distilling Closed-Source LLM's Knowledge for Locally Stable and Economic Biomedical Entity Linking}
%
%
\author{Yihao Ai\inst{1,2} \thanks{These authors contributed equally.}
 \and
Zhiyuan Ning\inst{1,2} $^\star$
 \and
Weiwei Dai\inst{3} \and
Pengfei Wang\inst{1,2} \and
Yi Du\inst{1,2} \and
Wenjuan Cui\inst{1,2} \thanks{Corresponding Author} \and
Kunpeng Liu\inst{4}  $^{\star\star}$\and
Yuanchun Zhou\inst{1,2}
}
\authorrunning{Y.Ai et al.}
%
\institute{Computer Network Information Center, Chinese Academy of Sciences, Beijing, China \and
University of Chinese Academy of Sciences, Beijing, China \\
\email{wenjuancui@cnic.cn} \and
 Changsha Aier Eye Hospital, Changsha, China \and
 Portland State University, Oregon, United States \\
\email{kunpeng@pdx.edu}}
\maketitle              
\begin{abstract}
Biomedical entity linking aims to map nonstandard entities to standard entities in a knowledge base. Traditional supervised methods perform well but require extensive annotated data to transfer, limiting their usage in low-resource scenarios. Large language models (LLMs), especially closed-source LLMs, can address these but risk stability issues and high economic costs: using these models is restricted by commercial companies and brings significant economic costs when dealing with large amounts of data. To address this, we propose ``RPDR'', a framework combining closed-source LLMs and open-source LLMs for re-ranking candidates retrieved by a retriever fine-tuned with a small amount of data. By prompting a closed-source LLM to generate training data from unannotated data and fine-tuning an open-source LLM for re-ranking, we effectively distill the knowledge to the open-source LLM that can be deployed locally, thus avoiding the stability issues and the problem of high economic costs. We evaluate RPDR on two datasets, including one real-world dataset and one publicly available dataset involving two languages: Chinese and English. RPDR achieves 0.019 Acc@1 improvement and 0.036 Acc@1 improvement on the Aier dataset and the Ask A Patient dataset when the amount of training data is not enough. The results demonstrate the superiority and generalizability of the proposed framework.

\keywords{Entity Linking  \and  LLM \and Prompt Engineering \and Knowledge Distillation.}
\end{abstract}
\section{Introduction}
Biomedical texts, such as medical literature and electronic medical records~\cite{ye2023needed}, contain valuable information. However, due to different standards and varied writing habits, nonstandard medical entities (called ``mention'') are commonly present in biomedical~\cite{xu2024sccdcg,xu2025scsiameseclu,ning2025deep} texts, posing challenges in utilizing this information. Therefore, it is important to perform biomedical entity linking, which maps a mention to an entity in a knowledge base~\cite{dong2024temporal,dong2023adaptive,qiao2020context,ning2021lightcake,dong2025disentangled} which consists of a large number of standard entities.

Previous biomedical entity linking methods were mainly based on supervised learning, often focusing on specific disciplines and languages~\cite{cnn,cmtn}. However, due to the differences among disciplines and languages, the trained models are not able to link mentions in other disciplines or languages within the biomedical domain, thus requiring lots of human-labeled data to retrain the models, which makes the methods impractical for new situations or tasks where pre-existing labeled data is limited. We refer to the above problem as ``\textbf{transfer distortion}''. To handle low-resource scenarios, methods with few-shot capabilities are urgently needed.

Recently, large language models have shown promising power in various tasks. LLMs like GPT-3.5 Turbo and GPT-4~\cite{gpt4} can easily transfer across domains and languages with just several examples~\cite{cai2023resolving}, making them promising for addressing the transfer distortion problem in biomedical entity linking. However, the best-performing large language models are either closed-source models or open-source models with massive parameter sizes, which have many limitations in practical applications. These models are typically accessed via commercial APIs, making their stability highly dependent on the companies' server status. And if a company discontinues its API service, related tasks become impossible to perform. Moreover, API usage incurs significant costs, especially given vast volume of medical text requiring linking. We refer to the above challenge as ``\textbf{costs and stability issues}''. Open-source LLMs such as Llama~\cite{llama} with smaller parameter sizes avoid these issues but may not be powerful enough to solve the task due to limited inference ability. Based on the above statements, we propose a question: \textbf{\textit{Can we build a framework that easily transfers between different disciplines and languages while addressing the costs and stability issues?}}

As closed-source LLMs demonstrate strong transfer capabilities and open-source LLMs avoid high costs and stability concerns, we combine closed-source and open-source LLMs by distilling knowledge from closed-source LLMs to open-source LLMs, obtaining a powerful LLM that can be deployed locally and avoid the problems.

Previous works~\cite{xu2020generate} mainly used a two-step framework of ``candidate retrieval and candidate re-ranking''. In this paper, we reformulate this into a new three-step framework: candidate \textbf{R}etrieval, \textbf{P}rompting-based training data generation, and \textbf{D}istillation for candidate \textbf{R}e-ranking (\textbf{RPDR}). For candidate retrieval, we apply the retriever proposed by Xu et al.~\cite{improving} and fine-tune it with limited human-labeled data. For candidate re-ranking, we divide the process into two sub-steps. First, we design a prompt for closed-source LLMs to annotate the data, reducing annotation costs with LLM's few-shot ability. Subsequently, with the generated data, the open-source LLMs are fine-tuned and imitate the closed-source LLMs' output, obtaining knowledge from closed-source LLMs. With knowledge distillation, a locally deployed open-source model can obtain the closed-source model's ability in biomedical entity linking, thus avoiding continuously calling APIs to use LLMs. The framework can transfer across disciplines and languages as closed-source LLMs can understand mentions in various languages and disciplines and can generate data in the required formats.

In summary, we propose a three-step framework named RPDR for biomedical entity linking. When training data is not enough, with prompt engineering, we achieve performance improvement on two datasets. With the complete framework, the performance is further improved. The main contributions of this paper are as follows: (1) We redesign the traditional two-step framework into a three-step framework and achieve performance improvement on two datasets, showing the effectiveness of the proposed framework. (2) We are the first to apply knowledge distillation in biomedical entity linking, taking advantage of a closed-source LLM to generate training data and fine-tune an open-source model thus obtaining a locally deployed model. (3) We carry out experiments on real-world and public datasets to validate the effectiveness of our method.

\section{Related Work}

\subsection{Biomedical Entity Linking}

Recent biomedical entity linking methods mainly adopt a ``candidate retrieval and candidate re-ranking'' framework. Li et al.~\cite{cnn} combined rule-based retrieval with CNNs to measure mention-entity similarities. Ji et al.~\cite{bertbased} enhanced candidate re-ranking using pre-trained BERT, BioBERT, and ClinicalBERT. Liu et al.~\cite{sapbert} proposed SapBERT, improving entity representation via contrastive learning. Xu et al.~\cite{mtcen} re-ranked the candidates only once with all candidates in one prompt, reducing prediction errors.

\subsection{Prompt Engineering}

With the rise of LLMs like ChatGPT, people have started designing prompts to leverage LLMs' capabilities in various tasks. Wei et al.~\cite{CoT} introduced Chain of Thought, breaking tasks into reasoning steps, achieving remarkable performance. Yao et al.~\cite{ToT} proposed Tree of Thoughts, allowing LLMs to consider multiple different reasoning paths and perform self-evaluation. Prompt engineering is also applied in temporal relation extraction~\cite{temporalrelation}, authorship verification~\cite{wrote}, and more.

\subsection{Knowledge Distillation}

Knowledge distillation distills knowledge from a large teacher model to a smaller student model. The student model can efficiently learn from a teacher model by mimicking the teacher model's output~\cite{knowledgedistillation}. West et al.~\cite{symbolic} distilled causal commonsense from GPT-3 to a small model, resulting in a compact commonsense model. Wang et al.~\cite{selfinstruct} introduced SELF-INSTRUCT, a framework where a pre-trained model generates its own training data, which is then filtered and used to fine-tune the model.

\begin{figure*}[!t]
\includegraphics[width=1.0\linewidth]{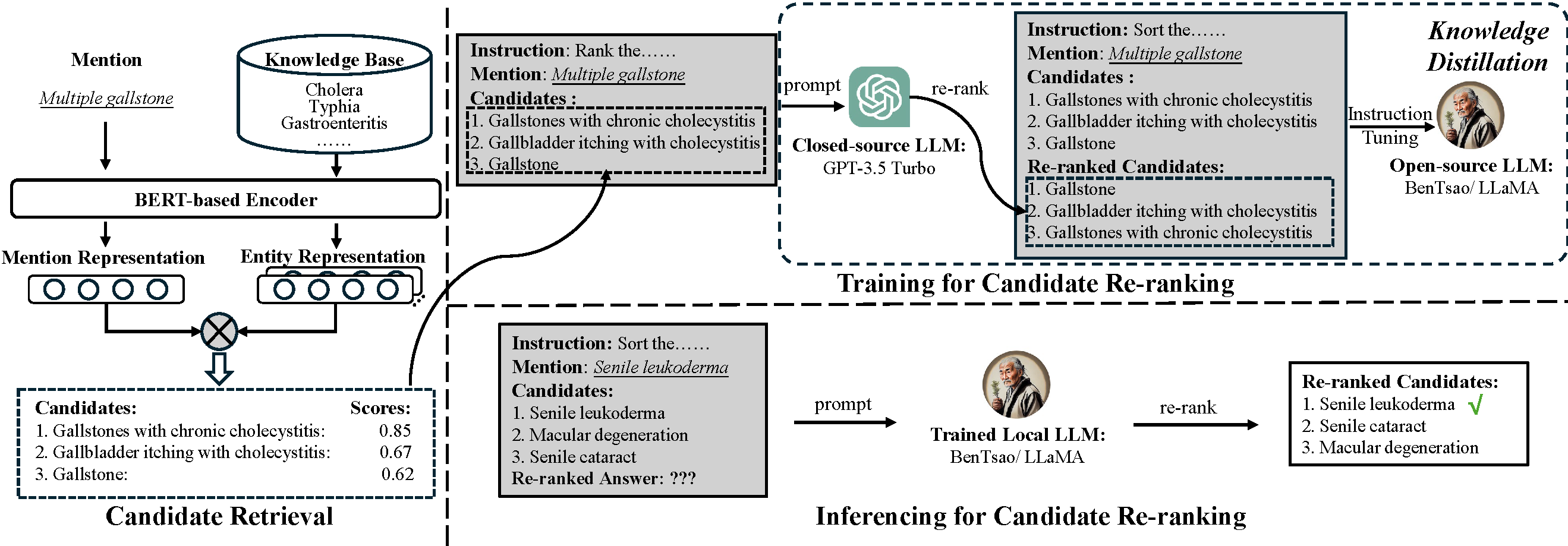}
\caption{Overall framework of our method. Candidates are generated by a bi-encoder. In the training step, the candidates are re-ranked by closed-source LLMs and used for training open-source LLMs. When inferencing, the candidates are re-ranked by the fine-tuned specialized LLMs.}
\label{fig:framework}
\end{figure*}

\section{Method}
\textbf{\textit{Biomedical entity linking}} aims to map nonstandard entities (called ``mention'') to corresponding standard entities (called ``golden entity'') in a knowledge base. Formally, given a mention $m$ with context $c$ in biomedical text and a knowledge base $\varepsilon$ containing $N$ entities ($\varepsilon=(e_1,e_2,...,e_N)$), the task is to find the corresponding golden entity $e_i$ that has the same meaning as $m$.

Many of the previous work adopted a two-step framework (candidate retrieval and candidate re-ranking) to handle biomedical entity linking. We rethink the task and propose a new three-step framework (as Fig.~1 shows), namely ``candidate retrieval, prompting-based training data generation and distillation for candidate re-ranking''.
\begin{enumerate}
    \item \textbf{Candidate Retrieval}: Retrieving $k$ candidates from the knowledge base.
    \item \textbf{Prompting-based Training Data Generation}: Prompting a closed-source LLM to re-rank the candidates, generating training data for the next step.
    \item \textbf{Distillation for Candidate Re-ranking}: Distilling knowledge from a closed-source LLM to an open-source model, using the data generated by the last step to fine-tune the open-source model.
\end{enumerate}

\subsection{Candidate Retrieval}
We directly adopt the method proposed by Xu et al.~\cite{improving}, employing a bi-encoder initiated from a BERT-based model to obtain the representations of the mention and all entities in the knowledge base and then compute their similarity using inner product scoring. Entities with the highest similarity scores are selected as candidates.

The representations are obtained using the following formula:

\begin{equation}
\begin{gathered}
r_m = \text{BERT}(\text{mention + context}), \\
r_e = \text{BERT}(\text{entity}),
\end{gathered}
\end{equation}
where $r_m$ and $r_e$ are the representations of the mention and an entity in the knowledge base, $BERT$ refers to the BERT-based model we use, and $context$ represents the context of the mention. The last hidden state of $[CLS]$ token is used as the representation of the mention and the entity. During training, entities highly similar to the mention but not being the golden entity are selected as hard negatives. These, along with randomly selected entities, are combined with the mentions to construct negative sample pairs.

For English datasets, we directly utilize the pretrained SapBERT to obtain the representationss for mentions and entities without any fine-tuning. The representations are derived directly from the mention text and the entity text. Then, the inner product is applied to compute the score of the (mention, entity) pair:

\begin{equation}
    score(mention,entity)={r_m} \cdot {r_e}.
\end{equation}

During inference, candidates are selected by computing the inner products, and entities with the highest scores are passed to the next step.

\subsection{Prompting-based Training Data Generation}

Previous work~\cite{cmtn,mtcen} mainly used cross-encoders for re-ranking, requiring substantial human-labeled data. We generate training data for the open-source LLMs using the few-shot learning ability of the closed-source LLMs. The process of generating training data using closed-source LLMs is actually the process of candidate re-ranking performed by the same LLM.

Given a mention $m$ and $k$ candidates $(c_1, c_2,..., c_k)$ obtained in the candidate retrieval step, the re-ranking~\cite{ning2022graph,ning2025rethinking} task aims to rank the candidates according to their probabilities of being the golden entity to which the mention refers. The re-ranked candidates are represented as $(e_1, e_2,..., e_k)$, where $e_1$ is the candidate with the highest probability of being the golden entity and $e_k$ with the lowest probability.

To guide the LLM in understanding and completing the re-ranking task, we employ prompt engineering. The designed prompt is shown in Fig.~2, where mention should be replaced by a real mention, and ``candidate1, candidate2,..., candidate6'' represents the candidates. In the prompt, we first instruct the closed-source LLM to understand the mention and then specify the task it needs to complete. By defining the output format and providing examples, we guide the LLM to produce the desired result.

\begin{figure}[!t]
  \centering
  \includegraphics[width=\linewidth]{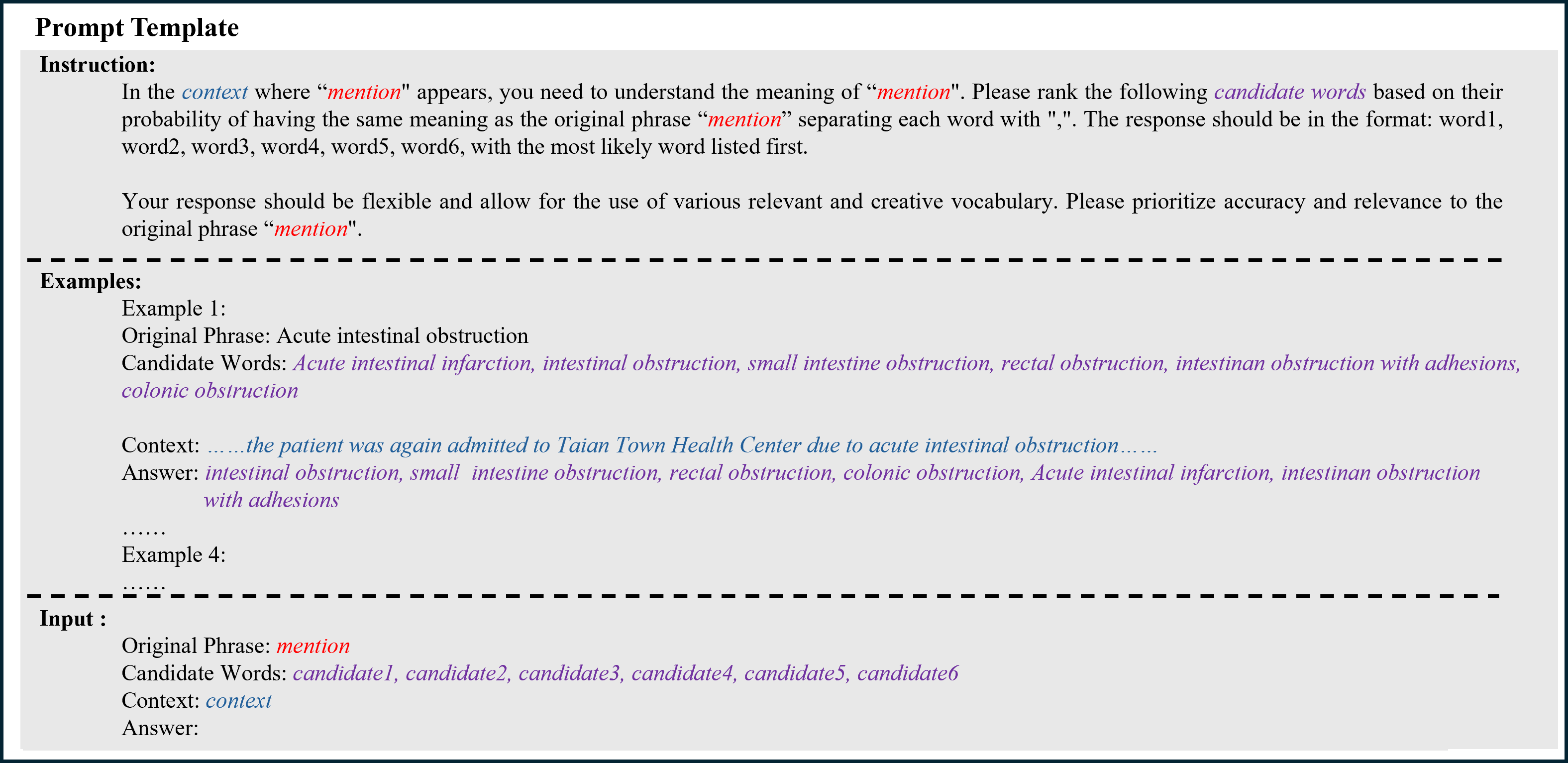}
  \caption{Illustration of the designed prompt.}
  \label{fig: promptExample}
\end{figure}

Formally, the overall process for candidate re-ranking of an LLM is as follows:

\begin{equation}
    rec=LLM(prompt, query),
\end{equation}
where $query$ consists of the mention, the candidates, and the context of the mention.
And $rec$ is the output of the LLM, consisting of $k$ re-ranked candidates. With closed-source LLMs, we obtain a batch of mentions paired with re-ranked candidates, along with corresponding candidates and contexts, these are then used to fine-tune the open-source LLMs.

\subsection{Distillation for Candidate Re-ranking}
Fine-tuning open-source large language models with domain-specific data can enhance their capabilities, while annotating such data is labor-intensive and costly. We use the data depicted in section 3.2 to fine-tune an open-source model for the re-ranking task, distilling the re-ranking ability from the closed-source LLM into the open-source LLM.

With the mentions, corresponding candidates and re-ranked candidates, we design a prompt template and fill it with mentions and corresponding candidates, guiding the open-source LLMs to learn the orders of re-ranked candidates. After fine-tuning, the open-source LLMs learn to re-rank and output candidates in the same way as the closed-source LLM. The process can be formulated as follows:

\begin{equation}
\begin{aligned}
   \theta = &\mathop{\arg\max}\limits_{\theta} \ p(entities \mid \\
   &LLM, instruction, mention_i, candidates_i; \theta_0, \theta) ,
\end{aligned}
\end{equation}
here, $entities$ are re-ranked candidates generated by closed-source LLMs. The instruction used for training open-source LLMs is shown in Fig.~3. We specify the task in the instruction, then mention and candidates are added in the instruction. The output format or examples are not needed. The open-source LLMs are fine-tuned with the Low-Rank Adaption (LoRA) method~\cite{lora}, $\theta_0$ represents the parameters of the pre-trained models, $\theta$ represents the trainable low-rank parameter increment.

During inference, we first retrieve several entities from the knowledge base as candidates according to the similarity between the mention and the entities, then the candidates are re-ranked by the fine-tuned open-source LLM, without any participation from the closed-source LLM.

\begin{figure}[!t]
  \centering
  \includegraphics[width=\linewidth]{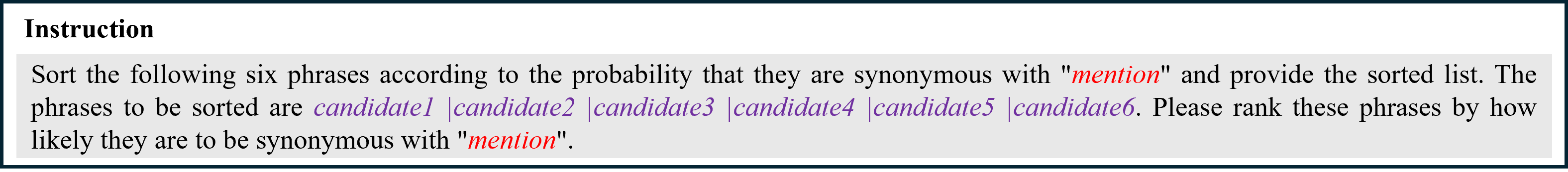}
  \caption{Illustration of the designed instruction used for fine-tuning BenTsao.}
  \label{fig: instruction}
\end{figure}

\section{Experiments}

\subsection{Datasets}
For evaluation, we utilize two datasets, with statistics summarized in Table~1. \textbf{The Aier Dataset} is a Chinese medical domain-specific corpus for entity linking. Mentions are extracted from electronic clinical records and annotated by medical professionals, with ICD-10 serving as the reference knowledge base. \textbf{The Ask A Patient Dataset}~\cite{askapatient} is an English biomedical entity linking dataset, containing mentions related to adverse drug reactions. Mentions are extracted from askapatient.com blog posts, annotated with two knowledge bases, including SNOMED-CT and AMT. For this dataset, we use the first fold for evaluation.

\noindent \textbf{Acknowledgment}: The Aier dataset is a private dataset with full authorization from Aier Eye Hospital Group Company Limited and has been desensitized for privacy preservation. The Ask A Patient dataset is publicly available.

\begin{table}[!t]
\caption{ Statistics of the Aier Dataset and the Ask A Patient Dataset.}
\begin{center}
\setlength{\tabcolsep}{3mm}
{
\begin{tabular}{lrrr}
\hline
\textbf{Dataset} & \textbf{Train} & \textbf{Val} & \textbf{Test}\\
\hline
Aier  & 309& 103 &104\\
Ask A Patient & 15612 & 845 & 867\\

\hline
\end{tabular}}
\label{tab：freq}
\end{center}
\end{table}

\subsection{Evaluation Metrics}
We assess performance via Acc@$k$ metrics for $k \in \{1, 5\}$, where Acc@$k$ represents the proportion of instances with the golden entity ranked among the top-$k$ model predictions. Concretely, a prediction is considered correct under Acc@$k$ if the entity exactly matching the mention is included within the highest $k$ re-ranked candidates.

\subsection{Baselines}
We compare our proposed framework against seven baselines: TF-IDF is a statistical method that considers the frequencies of a word's appearance in a certain document and its frequency across all the documents. Sieve-based~\cite{sieve} is a rule-based method, which designs ten rules to filter the entities and find the golden entity. BioSyn~\cite{sung2020biomedical} generates sparse representations~\cite{ning2024fedgcs} and dense representations for mentions and candidates. BioSyn is trained with hard negatives. SapBERT~\cite{sapbert} narrows the distance between the representations for entities with the same CUI. BioCEI~\cite{improving} designs a masked prompt with all the candidates present in the prompt so that the model can see all the candidates. CMTN~\cite{cmtn} designs a keyword attentive re-ranker to pay attention to the special parts of the mentions and the candidates. MTCEN~\cite{mtcen} proposes a multi-task framework, optimizing its re-ranker by correctly predicting the golden entity and the implication number.

For these baselines, when evaluated on the Aier dataset where synonyms are unavailable, we omitted modules reliant on synonym information. For methods leveraging SapBERT or other biomedical text-pretrained models, which lack direct Chinese biomedical counterparts, we replaced them with the Chinese-roberta-wwm-ext-large model as an alternative. When evaluated on the Ask A Patient dataset, the results of the baselines are reported without fine-tuning. Specifically, the result for SapBERT is reported using the pre-trained SapBERT directly. The results for BioSyn and BioCEI are obtained using checkpoints fine-tuned on the NCBI-Disease dataset. These checkpoints are applied to the Ask A Patient dataset without further fine-tuning, simulating a scenario where no training data is available for the target dataset.

\subsection{Implementation Details}
For candidate retrieval, the retriever implementation on the Aier dataset adheres to the training strategies and default hyperparameter configurations outlined in~\cite{improving}. Specifically, learning rate is set to 2e-6, weight decay to 0.01, batch size to 2, maximum context length to 256, candidate number to 6, and the number of negatives to 15, 10\% of which are hard negatives. Specially, for the Aier dataset, training spans 40 epochs, with the Chinese-roberta-wwm-ext-large model serving as the backbone encoder. For the Ask A Patient dataset, SapBERT is used without any fine-tuning. The max input length and batch size for SapBERT are set to 25 and 16 separately.

For prompting-based training data generation, we use GPT-3.5-Turbo-0125 (referred to as GPT-3.5 Turbo) and DeepSeek V3, which demonstrates performance comparable to the most powerful closed-source LLMs, to generate re-ranking candidates responses for the subsequent third step on the Aier and Ask A Patient datasets separately, and the number of examples in the prompt is 4 and 5 separately. The temperature is fixed at 0 during generation for the two models.

For distillation for candidate re-ranking, we leverage BenTsao~\cite{huatuo} on the Aier dataset, which is a model fine-tuned on the Chinese medical knowledge graph. For the Ask A Patient dataset, we leverage Llama2-7B (referred to as Llama2)~\cite{llama}. The training data comprise 1000 and 2000 generated samples separately. The models are fine-tuned via LoRA, a low-rank adaptation technique~\cite{lora}. We mainly use the default hyperparameters in~\cite{huatuo} to fine-tune BenTsao. Specifically, learning rate is set to 3e-4, batch size to 8, epoch to 10, LoRA rank to 8, LoRA alpha to 16, LoRA dropout to 0.05. When fine-tuning Llama-2, learning rate is set to 3e-5, epoch to 5, batch size to 8, with other hyper-parameters the same. All the models are fine-tuned on an NVIDIA A100 GPU.

\subsection{Overall Performance}
We evaluate our method against seven baselines depicted in section 4.3, with results listed in Table~2. By just prompting raw LLMs, namely GPT-3.5 Turbo and DeepSeek V3 on the two datasets, we outperform all the baselines on both datasets. With the complete framework, performance improves further.

On the Aier dataset, using GPT-3.5 Turbo, we achieve 0.01 Acc@1 improvement compared to all the baselines. With the complete framework, we further achieve 0.009 Acc@1 improvement and 0.029 Acc@5 improvement compared to using GPT-3.5 Turbo directly. On the Ask A Patient dataset, when training data is unavailable, supervised learning-based methods such as BioSyn and BioCEI perform poorly and may not even surpass rule-based approach. Our method beats all the baselines when using DeepSeek V3 for re-ranking, achieving 0.026 Acc@1 and 0.005 Acc@5 improvement. When using the complete framework, Acc@1 improves again, achieving 0.734 Acc@1, showing the effectiveness of our method.

\begin{table*}[!t]
  \centering
  \caption{Results of our method and the baselines. The best result of each metric is highlighted in bold. “-” means not reproduced on the dataset or not computed in the provided code.}
  \label{tab:overallperformance}
  \setlength{\tabcolsep}{3mm}
  \begin{tabular}{lcccc}
\hline
\multirow{2}{*}{\textbf{Model}} & \multicolumn{2}{c}{\textbf{Aier}} &  \multicolumn{2}{c}{\textbf{Ask A Patient}}\\

\cline{2-3}
\cline{4-5}

 & Acc@1 & Acc@5 & Acc@1 & Acc@5 \\
\hline
TF-IDF & 0.462 & 0.519 & 0.263 & 0.276 \\
Sieve-based & 0.279 & - & 0.592 & - \\
BioSyn & 0.702 & 0.859 & 0.559 & 0.759 \\
SAPBERT & 0.673 & 0.798 & 0.698 & 0.875 \\
BioCEI & 0.721 & 0.904 & 0.295 & 0.311 \\
CMTN & 0.721 & 0.904 & -  & - \\ 
MTCEN & 0.682 &0.834 & - & - \\
\hline
Raw LLM & 0.731 & 0.904 & 0.724 & \textbf{0.880}  \\
Our Method & \textbf{0.740} & \textbf{0.933} & \textbf{0.734} & 0.871 \\
\hline
\end{tabular}
\end{table*}

\subsection{Ablation Studies}
We study the impact of LLMs for candidate re-ranking with the same prompt. As shown in Table~3, more powerful LLMs like GPT-4 demonstrate superior performance due to better reasoning capabilities while Llama-2-7B and Chinese-Alpaca-2-7B~\cite{Chinese-LLaMA-Alpaca} without fine-tuning perform badly, reflecting that small open-source LLMs are not powerful enough to conduct re-ranking directly.

\begin{table}[!t]
\centering
\caption{Performances comparison between different LLMs with the same prompt. Llama-2 represents Llama-2-7B. Alpaca-2 represents Chinese-Alpaca-2-7B.}
\label{tab:ablationofprompt}
\setlength{\tabcolsep}{3mm}
\begin{tabular}{lcccc}
\hline
\multicolumn{2}{l}{\textbf{Dataset}} & \textbf{LLM} & \textbf{Acc@1} & \textbf{Acc@5}\\
\hline
\multicolumn{2}{l}{\multirow{4}{*}{Aier}} & \multicolumn{1}{c}{GPT-3.5 turbo}  & 0.731 & 0.904\\
\multicolumn{2}{l}{}& GPT-4 & \textbf{0.808} & \textbf{0.933} \\
\multicolumn{2}{l}{}& Llama-2 & 0.452 & 0.702 \\
\multicolumn{2}{l}{}& Alpaca-2 & 0.231 & 0.298 \\
\hline
\end{tabular}
\end{table}

We compare the performance of open-source LLMs fine-tuned using human-labeled data versus GPT-3.5 Turbo-generated data. Table~4 shows that using 1000 mention-candidate pairs generated by GPT-3.5 Turbo results in improved performance over models fine-tuned solely with human-annotated data. This indicates that our method can perform well even without human-annotated data.

\begin{table}[!t]
\centering
\caption{Performance comparison between open-source LLMs fine-tuned with human-labeled data and GPT-3.5 Turbo generated data.}
\label{tab:training_data}
\setlength{\tabcolsep}{3mm}
\begin{tabular}{lcccc}
\hline
\multicolumn{2}{l}{\textbf{Dataset}} & \multicolumn{2}{c}{\textbf{Training Data}} & \textbf{Acc@1} \\
\hline
\multicolumn{2}{l}{\multirow{2}{*}{Aier}} & \multicolumn{2}{c}{Human-labeled Data}  & 0.712\\
\multicolumn{2}{l}{}& \multicolumn{2}{c}{GPT-3.5 Turbo Generated Data} & \textbf{0.740} \\

\hline
\end{tabular}
\end{table}

We assess the impact of generated training data sizes on the performance of fine-tuned open-source LLM. As shown in Fig.~4, as the number of generated data increases, the performance initially improves, after reaching the upper peak, the performance drops, which may be because that with the increment of the number of the training data, the LLM is fine-tuned more fully thus performs better, but then with the number increases continually, more wrong mention-candidates pairs are introduced and hurt the performance of the LLM.

\begin{figure}[!t]
  \centering
  \includegraphics[width=0.70\linewidth]{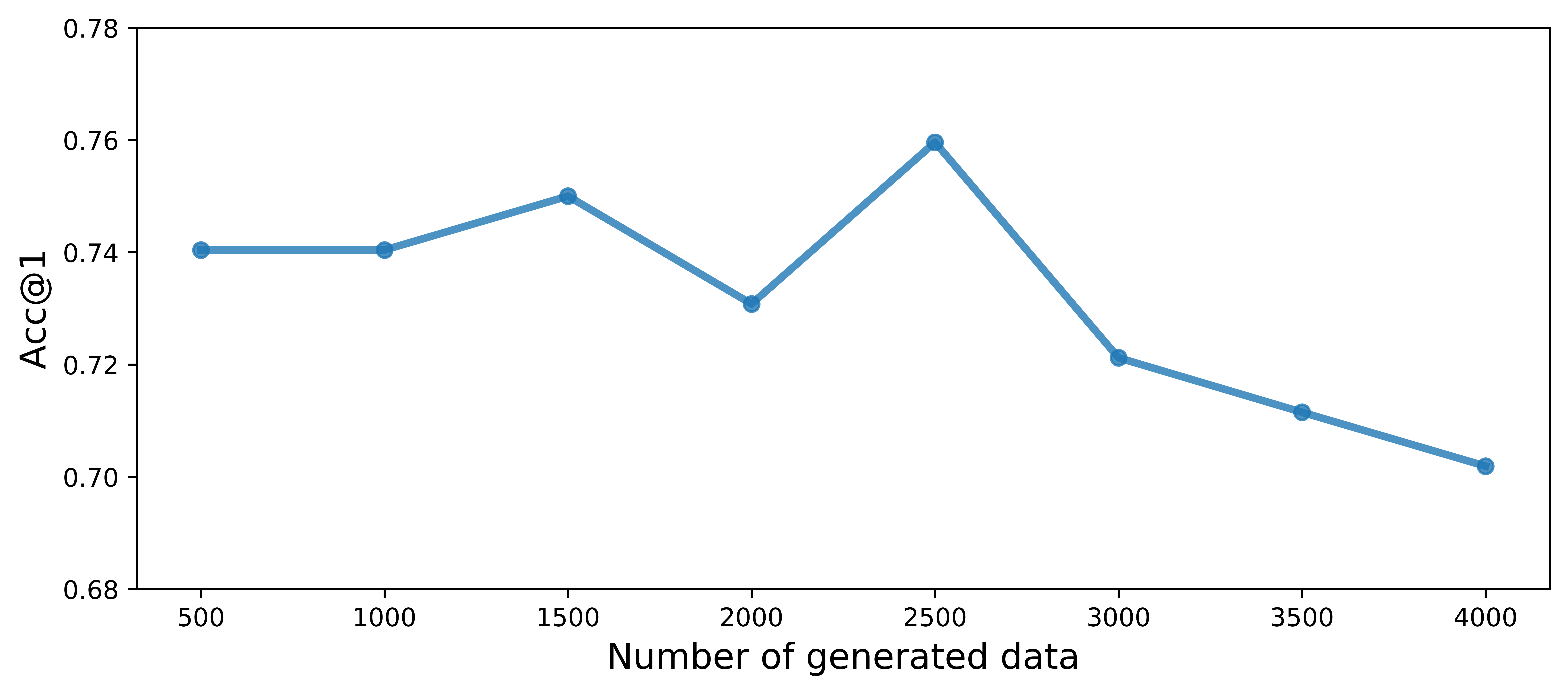}
  \caption{The performances of the fine-tuned LLM with different numbers of training data.}
  \label{fig: number_of_candidates}
\end{figure}

To study the effects of medical knowledge, we compare BenTsao's performance with three LLMs without medical knowledge: Llama2-7B, Alpaca-2 and Huozi~\cite{Chinese-LLaMA-Alpaca}, the base model of BenTsao. As Table~5 shows, on the Aier dataset, BenTsao performs better than the other three LLMs, showing that including medical knowledge may be helpful for understanding the mentions and entities, resulting in superior performance.

\begin{table}[!t]
\centering
\caption{Performance comparison between different base models used for fine-tuning on the Aier dataset, Llama-2 represents Llama-2-7B. Alpaca-2 represents Chinese-Alpaca-2-7B.}
\setlength{\tabcolsep}{5mm}{
\begin{tabular}{lcc}
\hline
\textbf{Model}&\textbf{Acc@1}&\textbf{Acc@5}\\
\hline
Huozi & 0.712 & 0.894\\
Llama-2 &0.712 & 0.913 \\
Alpaca-2 &0.702 & 0.923\\
BenTsao & \textbf{0.740} & \textbf{0.933}\\
\hline
\end{tabular}}
\end{table}

Finally, we compared the economic costs of using GPT-3.5 turbo, DeepSeek V3 and a locally deployed fine-tuned Llama2-7B for re-ranking on the Ask A Patient dataset. As Table~6 shows, deploying Llama2 locally significantly reduces inference costs. The costs for the locally deployed model are calculated based on GPU server inference time.

\begin{table}[!t]
\centering
\caption{Economic costs comparison between different models.}
\setlength{\tabcolsep}{5mm}{
\begin{tabular}{lccc}
\hline
\textbf{Model} & \textbf{GPT-3.5 turbo} & \textbf{DeepSeek V3} & \textbf{Local Llama-2}\\
\hline
Cost (\$) & 0.294 & 0.060 & 0.038 \\

\hline
\end{tabular}}
\end{table}

\subsection{Conclusion}
In this paper, we propose a framework ``RPDR'' for biomedical entity linking, which combines closed-source LLMs and open-source LLMs, distilling the knowledge from closed-source LLMs to open-source LLMs. With only several examples, we prompt closed-source LLMs to generate training data and use the data to fine-tune open-source LLMs, eliminating the need for extensive human-labeled data and avoiding the transfer distortion problem. With the locally deployed open-source model, we avoid continuous reliance on closed-source large language models, thus avoiding high costs and stability issues. Results indicate that our method performs well on two datasets, demonstrating its effectiveness.

\begin{credits}
\subsubsection{\ackname} This work was supported by the Natural Science Foundation of China under Grant No. T2322027 and 62332204, Youth Innovation Promotion Association CAS. We thank all the doctors of Aier eye hospital who help to annotate and verify the corpus.

\end{credits}

%
%
%
%
\bibliographystyle{splncs04}  
\bibliography{reference.bib}  

\begin{thebibliography}{10}
\providecommand{\url}[1]{\texttt{#1}}
\providecommand{\urlprefix}{URL }
\providecommand{\doi}[1]{https://doi.org/#1}

\bibitem{gpt4}
Achiam, J., Adler, S., Agarwal, S., Ahmad, L., Akkaya, I., Aleman, F.L., Almeida, D., Altenschmidt, J., Altman, S., Anadkat, S., et~al.: Gpt-4 technical report. arXiv preprint arXiv:2303.08774  (2023)

\bibitem{cai2023resolving}
Cai, X., Xiao, M., Ning, Z., Zhou, Y.: Resolving the imbalance issue in hierarchical disciplinary topic inference via llm-based data augmentation. In: 2023 IEEE International Conference on Data Mining Workshops (ICDMW). pp. 1424--1429. IEEE (2023)

\bibitem{Chinese-LLaMA-Alpaca}
Cui, Y., Yang, Z., Yao, X.: Efficient and effective text encoding for chinese llama and alpaca. arXiv preprint arXiv:2304.08177  (2023), \url{https://arxiv.org/abs/2304.08177}

\bibitem{dong2023adaptive}
Dong, H., Ning, Z., Wang, P., Qiao, Z., Wang, P., Zhou, Y., Fu, Y.: Adaptive path-memory network for temporal knowledge graph reasoning. arXiv preprint arXiv:2304.12604  (2023)

\bibitem{dong2025disentangled}
Dong, H., Qiao, Z., Ning, Z., Hao, Q., Du, Y., Wang, P., Zhou, Y.: Disentangled multi-span evolutionary network against temporal knowledge graph reasoning. arXiv preprint arXiv:2505.14020  (2025)

\bibitem{dong2024temporal}
Dong, H., Wang, P., Xiao, M., Ning, Z., Wang, P., Zhou, Y.: Temporal inductive path neural network for temporal knowledge graph reasoning. Artificial Intelligence  \textbf{329},  104085 (2024)

\bibitem{sieve}
D’Souza, J., Ng, V.: Sieve-based entity linking for the biomedical domain. In: Proceedings of the 53rd Annual Meeting of the Association for Computational Linguistics and the 7th International Joint Conference on Natural Language Processing (Volume 2: Short Papers). pp. 297--302 (2015)

\bibitem{knowledgedistillation}
Gou, J., Yu, B., Maybank, S.J., Tao, D.: Knowledge distillation: A survey. International Journal of Computer Vision  \textbf{129}(6),  1789--1819 (2021)

\bibitem{lora}
Hu, E.J., Wallis, P., Allen-Zhu, Z., Li, Y., Wang, S., Wang, L., Chen, W., et~al.: Lora: Low-rank adaptation of large language models. In: International Conference on Learning Representations (2021)

\bibitem{wrote}
Hung, C.Y., Hu, Z., Hu, Y., Lee, R.K.W.: Who wrote it and why? prompting large-language models for authorship verification. In: The 2023 Conference on Empirical Methods in Natural Language Processing (2023)

\bibitem{bertbased}
Ji, Z., Wei, Q., Xu, H.: Bert-based ranking for biomedical entity normalization. AMIA Summits on Translational Science Proceedings  \textbf{2020}, ~269 (2020)

\bibitem{cnn}
Li, H., Chen, Q., Tang, B., Wang, X., Xu, H., Wang, B., Huang, D.: Cnn-based ranking for biomedical entity normalization. BMC bioinformatics  \textbf{18},  79--86 (2017)

\bibitem{cmtn}
Liang, M., Xue, K., Ye, Q., Ruan, T.: A combined recall and rank framework with online negative sampling for chinese procedure terminology normalization. Bioinformatics  \textbf{37}(20),  3610--3617 (2021)

\bibitem{askapatient}
Limsopatham, N., Collier, N.: Normalising medical concepts in social media texts by learning semantic representation. In: Proceedings of the 54th Annual Meeting of the Association for Computational Linguistics (volume 1: long papers). pp. 1014--1023 (2016)

\bibitem{sapbert}
Liu, F., Shareghi, E., Meng, Z., Basaldella, M., Collier, N.: Self-alignment pretraining for biomedical entity representations. In: Proceedings of the 2021 Conference of the North American Chapter of the Association for Computational Linguistics: Human Language Technologies. pp. 4228--4238 (2021)

\bibitem{ning2021lightcake}
Ning, Z., Qiao, Z., Dong, H., Du, Y., Zhou, Y.: Lightcake: A lightweight framework for context-aware knowledge graph embedding. In: Pacific-Asia Conference on Knowledge Discovery and Data Mining. pp. 181--193. Springer (2021)

\bibitem{ning2024fedgcs}
Ning, Z., Tian, C., Xiao, M., Fan, W., Wang, P., Li, L., Wang, P., Zhou, Y.: Fedgcs: A generative framework for efficient client selection in federated learning via gradient-based optimization. arXiv preprint arXiv:2405.06312  (2024)

\bibitem{ning2025rethinking}
Ning, Z., Wang, P., Qiao, Z., Wang, P., Zhou, Y.: Rethinking graph contrastive learning through relative similarity preservation. arXiv preprint arXiv:2505.05533  (2025)

\bibitem{ning2022graph}
Ning, Z., Wang, P., Wang, P., Qiao, Z., Fan, W., Zhang, D., Du, Y., Zhou, Y.: Graph soft-contrastive learning via neighborhood ranking. arXiv preprint arXiv:2209.13964  (2022)

\bibitem{ning2025deep}
Ning, Z., Wang, Z., Zhang, R., Xu, P., Liu, K., Wang, P., Ju, W., Wang, P., Zhou, Y., Cambria, E., et~al.: Deep cut-informed graph embedding and clustering. arXiv preprint arXiv:2503.06635  (2025)

\bibitem{qiao2020context}
Qiao, Z., Ning, Z., Du, Y., Zhou, Y.: Context-enhanced entity and relation embedding for knowledge graph completion. arXiv preprint arXiv:2012.07011  (2020)

\bibitem{mtcen}
Sui, X., Song, K., Zhou, B., Zhang, Y., Yuan, X.: A multi-task learning framework for chinese medical procedure entity normalization. In: ICASSP 2022-2022 IEEE International Conference on Acoustics, Speech and Signal Processing (ICASSP). pp. 8337--8341. IEEE (2022)

\bibitem{sung2020biomedical}
Sung, M., Jeon, H., Lee, J., Kang, J.: Biomedical entity representations with synonym marginalization. In: Proceedings of the 58th Annual Meeting of the Association for Computational Linguistics. pp. 3641--3650 (2020)

\bibitem{llama}
Touvron, H., Lavril, T., Izacard, G., Martinet, X., Lachaux, M.A., Lacroix, T., Rozi{\`e}re, B., Goyal, N., Hambro, E., Azhar, F., et~al.: Llama: Open and efficient foundation language models. arXiv preprint arXiv:2302.13971  (2023)

\bibitem{huatuo}
Wang, H., Liu, C., Xi, N., Qiang, Z., Zhao, S., Qin, B., Liu, T.: Huatuo: Tuning llama model with chinese medical knowledge. arXiv preprint arXiv:2304.06975  (2023)

\bibitem{selfinstruct}
Wang, Y., Kordi, Y., Mishra, S., Liu, A., Smith, N.A., Khashabi, D., Hajishirzi, H.: Self-instruct: Aligning language models with self-generated instructions. In: Proceedings of the 61st Annual Meeting of the Association for Computational Linguistics (Volume 1: Long Papers). pp. 13484--13508 (2023)

\bibitem{CoT}
Wei, J., Wang, X., Schuurmans, D., Bosma, M., Xia, F., Chi, E., Le, Q.V., Zhou, D., et~al.: Chain-of-thought prompting elicits reasoning in large language models. Advances in neural information processing systems  \textbf{35},  24824--24837 (2022)

\bibitem{symbolic}
West, P., Bhagavatula, C., Hessel, J., Hwang, J., Jiang, L., Le~Bras, R., Lu, X., Welleck, S., Choi, Y.: Symbolic knowledge distillation: from general language models to commonsense models. In: Proceedings of the 2022 Conference of the North American Chapter of the Association for Computational Linguistics: Human Language Technologies. pp. 4602--4625 (2022)

\bibitem{xu2020generate}
Xu, D., Zhang, Z., Bethard, S.: A generate-and-rank framework with semantic type regularization for biomedical concept normalization. In: Proceedings of the 58th Annual Meeting of the Association for Computational Linguistics. pp. 8452--8464 (2020)

\bibitem{xu2025scsiameseclu}
Xu, P., Ning, Z., Li, P., Liu, W., Wang, P., Cui, J., Zhou, Y., Wang, P.: scsiameseclu: A siamese clustering framework for interpreting single-cell rna sequencing data. arXiv preprint arXiv:2505.12626  (2025)

\bibitem{xu2024sccdcg}
Xu, P., Ning, Z., Xiao, M., Feng, G., Li, X., Zhou, Y., Wang, P.: sccdcg: Efficient deep structural clustering for single-cell rna-seq via deep cut-informed graph embedding. In: International Conference on Database Systems for Advanced Applications. pp. 172--187. Springer (2024)

\bibitem{improving}
Xu, Z., Chen, Y., Hu, B.: Improving biomedical entity linking with cross-entity interaction. In: Proceedings of the AAAI Conference on Artificial Intelligence. vol.~37, pp. 13869--13877 (2023)

\bibitem{ToT}
Yao, S., Yu, D., Zhao, J., Shafran, I., Griffiths, T., Cao, Y., Narasimhan, K.: Tree of thoughts: Deliberate problem solving with large language models. Advances in Neural Information Processing Systems  \textbf{36} (2024)

\bibitem{ye2023needed}
Ye, X., Xiao, M., Ning, Z., Dai, W., Cui, W., Du, Y., Zhou, Y.: Needed: Introducing hierarchical transformer to eye diseases diagnosis. In: Proceedings of the 2023 SIAM International Conference on Data Mining (SDM). pp. 667--675. SIAM (2023)

\bibitem{temporalrelation}
Yuan, C., Xie, Q., Ananiadou, S.: Zero-shot temporal relation extraction with chatgpt. In: The 22nd Workshop on Biomedical Natural Language Processing and BioNLP Shared Tasks. pp. 92--102 (2023)

\end{thebibliography}
\end{document}